\documentclass[conference]{IEEEtran}
\IEEEoverridecommandlockouts
\usepackage{cite}
\usepackage{amsmath,amssymb,amsfonts}
\usepackage{algorithmic}
\usepackage{graphicx}
\graphicspath{{../figures/}{figures/}}
\usepackage{textcomp}
\usepackage{xcolor}
\usepackage{booktabs}
\usepackage{multirow}
\usepackage{placeins}
\usepackage[hidelinks]{hyperref}

\def\BibTeX{{\rm B\kern-.05em{\sc i\kern-.025em b}\kern-.08em
    T\kern-.1667em\lower.7ex\hbox{E}\kern-.125emX}}

\makeatletter
 \let\old@ps@headings\ps@headings
 \let\old@ps@IEEEtitlepagestyle\ps@IEEEtitlepagestyle
 \def\confheader#1{%
 \def\ps@IEEEtitlepagestyle{%
 \old@ps@IEEEtitlepagestyle%
 \def\@oddhead{\strut\hfill#1\hfill\strut}%
 \def\@evenhead{\strut\hfill#1\hfill\strut}%
 }%
 \ps@headings%
 }
 \newcommand{\linebreakand}{%
  \end{@IEEEauthorhalign}
  \hfill\mbox{}\par
  \mbox{}\hfill\begin{@IEEEauthorhalign}
}
 \makeatother

\confheader{%
 2026 International Conference on Intelligent Computing and Secure Communication Systems (ICICSCS)
 }

 \usepackage[pscoord]{eso-pic}
\newcommand{\placetextbox}[3]{
 \setbox0=\hbox{#3}
 \AddToShipoutPictureFG*{ \put(\LenToUnit{#1\paperwidth},\LenToUnit{#2\paperheight}){\vtop{{\null}\makebox[0pt][c]{#3}}}
 }
 }
 \placetextbox{.23}{0.055}{\small{979-8-3195-2455-3/26/\$31.00~\copyright 2026 IEEE}}

\usepackage{cite}
\usepackage{amsmath,amssymb,amsfonts}
\usepackage{algorithmic}
\usepackage{graphicx}
\usepackage{textcomp}
\def\BibTeX{{\rm B\kern-.05em{\sc i\kern-.025em b}\kern-.08em
    T\kern-.1667em\lower.7ex\hbox{E}\kern-.125emX}}

\begin{document}

\title{Promise and Limits of Cross-Architecture\\ Knowledge Distillation for Lightweight\\ Medical Vision Transformers (LiteMed-ViT)}
\author{\IEEEauthorblockN{Shuaibu Shehu}
\IEEEauthorblockA{\textit{School of CS \& AI, SR University} \\
Warangal, India-506371 \\
shuaibushehukalifa@gmail.com}
\and
\IEEEauthorblockN{Elijah M. Flomo}
\IEEEauthorblockA{\textit{School of CS \& AI, SR University} \\
Warangal, India-506371 \\
elijahmflomo@gmail.com}
\and
\IEEEauthorblockN{Mohammad Safiya}
\IEEEauthorblockA{\textit{Dept. of Computer Science, GITAM University} \\
Visakhapatnam, India-530045\\
mohammadsafiya1431@gmail.com}

\linebreakand 

\IEEEauthorblockN{Anurodh Kumar}
\IEEEauthorblockA{\textit{School of CS \& AI, SR University} \\
Warangal, India-506371 \\
anurodh.kumar@sru.edu.in}

\and
\IEEEauthorblockN{Subodh Kumar}
\IEEEauthorblockA{\textit{Dept. of Biochemistry} \\
\textit{Dr. KNS Memorial Institute of Medical Sciences} \\
Barabanki, India-225001 \\
somya2011.dsk@gmail.com}

}

\maketitle

\begin{abstract}
Running deep models for medical image analysis at the point of care is challenging with standard architectures: convolutional neural networks (CNNs) are often too heavy for edge devices, while Vision Transformers (ViTs) require massive training datasets to compensate for missing inductive biases. This study investigates how far the dual-token distillation used to train data-efficient ViTs can be pushed when applied across architectures, from a CNN-based teacher into a lightweight ViT student. LiteMed-ViT is introduced as a dual-token ViT featuring a convolutional stem and a class-balanced distillation objective; at 0.234M parameters, LiteMed-ViT carries 47.8$\times$ fewer parameters than its ResNet18 teacher at a memory footprint of just 0.89~MB. Evaluated against three compact CNN baselines on three MedMNIST subsets of varying modality and class balance, LiteMed-ViT attains higher accuracy than all three baselines on BloodMNIST (95.05\%) and two on PathMNIST (89.15\%). Under MedMNIST-C corruptions, it is at least as robust as every compact CNN across all subsets, outperforming them on BloodMNIST and PathMNIST without declining in relative standing as severity increases. Particular attention is devoted to identifying the exact conditions where performance plateaus. On DermaMNIST, LiteMed-ViT's balanced accuracy remains up to 1.9 points below the strongest compact CNNs and does not benefit from higher input resolutions up to 224 pixels, where ResNet18 gains 12.6 points. Stronger teachers do not yield a more accurate student, and overconfident ones lower its DermaMNIST balanced accuracy unless calibrated.
\end{abstract}


\section{Introduction}
Convolutional neural networks (CNNs) have driven much of the progress in medical image analysis in the past decade~\cite{litjens_survey}, but the point of care often lacks the memory and compute a standard CNN expects, and sending images to a server moves patient data off the device, adds latency and needs a reliable connection. The target is a model accurate enough to be useful and small enough to run locally~\cite{rancea_edge}.

Vision Transformers (ViTs) capture global context~\cite{dosovitskiy_vit} but lack convolutional locality and overfit small medical datasets. Knowledge distillation mitigates this by transferring a stronger model’s softened predictions~\cite{hinton_kd}, where a CNN teacher’s inductive bias is partially imparted to the student via DeiT’s distillation token~\cite{touvron_deit}.

This Study examines how far cross-architecture distillation can compress a ViT for medical image classification, and where its performance plateaus. The model, LiteMed-ViT, has 0.234M parameters and is distilled from ResNet18; it is evaluated on three MedMNIST v2 subsets~\cite{yang_medmnist} that differ in modality and class balance. Because differences between small models on these benchmarks are close to run-to-run variation, every model is trained with four seeds and compared under a pre-set significance rule and a matched decision rule (Section~\ref{sec:protocol}). The primary contributions are:
\begin{itemize}
    \item \textbf{LiteMed-ViT:} a 0.234M-parameter dual-token ViT with a convolutional stem and a class-balanced distillation objective, the smallest model, though not the fastest.
    \item \textbf{A four-seed benchmark against compact CNNs under matched decision rules:} LiteMed-ViT is more accurate than all three on BloodMNIST and two on PathMNIST, and up to 1.9 points below two on DermaMNIST.
   \item \textbf{Studies of the recipe's limits:} stronger teachers do not yield better students, overconfident teachers need calibration on imbalanced data, $T$ and $\alpha$ have little effect, and higher resolution does not help on DermaMNIST.
    \item \textbf{Robustness on \mbox{MedMNIST-C}:} LiteMed-ViT is at least as robust as every compact CNN on all subsets, and its relative standing does not decline with severity.
\end{itemize}

\section{Related Work}

\paragraph*{Compact architectures}
SqueezeNet matched AlexNet's accuracy with far fewer weights~\cite{iandola_squeezenet}, MobileNetV3 combined depthwise-separable convolutions with architecture search~\cite{howard_mobilenetv3}, and ShuffleNet~V2 argued that measured latency, not FLOPs, should guide design~\cite{ma_shufflenet}; MobileViT does the same for transformers~\cite{mehta_mobilevit}. The three convolutional networks serve as baselines.

\paragraph*{Vision Transformers and their data appetite}
Self-attention~\cite{vaswani_attention} was adapted to images by the ViT~\cite{dosovitskiy_vit}, which matches or exceeds CNNs given enough data, and the Swin Transformer reintroduced hierarchy and locality~\cite{liu_swin}. The dependence on large training sets is the recurring obstacle on medical data: comparisons of transformers and CNNs on medical tasks reach mixed conclusions~\cite{matsoukas_replace}, and surveys report rapid uptake alongside concerns about data efficiency~\cite{shamshad_survey}.

\paragraph*{Knowledge distillation}
Distillation trains a small network on the softened predictions of a larger one~\cite{hinton_kd}; later work matches intermediate features~\cite{romero_fitnets}, and a survey maps the variants~\cite{gou_survey}. DeiT adds a distillation token and finds a convolutional teacher effective~\cite{touvron_deit}. The distillation mechanism is applied at significantly higher compression ratios here, where a more accurate teacher is not necessarily better~\cite{cho_efficacy}. In medical imaging, hybrid CNN--transformer models classify blood cells~\cite{alshehri_hybrid} and, with dynamic dropout, chest X-rays~\cite{li_hycovit}; systematic investigations into the limits of cross-architecture distillation for ultra-lightweight models remain scarce.

\section{Methodology}

\subsection{Data and Augmentation}
\label{sec:data}

Three 2D subsets of MedMNIST~v2~\cite{yang_medmnist} are utilized: BloodMNIST (peripheral blood cells, 8 classes, $11{,}959$/$1{,}712$/$3{,}421$ train/val/test)~\cite{acevedo_blood}, DermaMNIST (dermatoscopic lesions, 7 classes, $7{,}007$/$1{,}003$/$2{,}005$)~\cite{tschandl_ham10000} and PathMNIST (colorectal histology, 9 classes, $89{,}996$/$10{,}004$/$7{,}180$)~\cite{kather_crc}, all $28\times28$ RGB. DermaMNIST is dominated by one class (predicting it alone yields $67\%$ accuracy); the other two are far more balanced (largest class $19.5\%$, $14.3\%$).

Because the three modalities vary in different ways, augmentation differs. BloodMNIST and DermaMNIST use geometry only (horizontal flips, rotations up to $15^\circ$), leaving colour, a diagnostic criterion in dermoscopy~\cite{nachbar_abcd}, undisturbed, although colour jitter can help melanoma classifiers~\cite{perez_aug}. PathMNIST's slides vary in stain intensity, and an unconstrained model can learn stain rather than tissue; therefore, reflect-padded crops are added, vertical flips and colour jitter ($0.3$ brightness/contrast/saturation, $0.05$ hue)~\cite{tellez_stain}. Inputs are normalised per channel with mean and standard deviation $0.5$.

Every choice of architecture and hyperparameter is made on the official validation split; each trained model is evaluated once on the test split.

Robustness to shifts in scanner, stain batch or illumination is measured using the released \mbox{MedMNIST-C} benchmark~\cite{disalvo_medmnistc} (v0.1.0): 11 modality-specific corruptions for BloodMNIST and PathMNIST and 15 for DermaMNIST, covering noise, blur, compression, colour and acquisition artefacts such as stain deposits, bubbles and dermatoscope black corners, each at five severities. The corrupted test sets were generated with the official code at $128\times128$ and resized to $28\times28$. The mean corrupted accuracy (mCA) over all corruptions and severities, as well as accuracy by severity, is reported; training never sees corrupted images.

\subsection{Teachers}
The teacher in every main experiment is a ResNet18~\cite{he_resnet} pretrained on ImageNet, fine-tuned on each subset at $28\times28$ for 50 epochs and then frozen. Its training loss is the cross-entropy \eqref{eq:ce_loss},

\begin{equation}
L_{CE}(z, y) = -\sum_{c=1}^{C} y_c \log\left(\sigma(z)_c\right),
\label{eq:ce_loss}
\end{equation}
where $C$ is the number of classes, $y$ the one-hot label and $\sigma(z)$ the softmax over the logits $z$. A teacher study replaces it with ResNet50~\cite{he_resnet}, \mbox{DenseNet-121}~\cite{huang_densenet} and the logit average of ResNet18 and ResNet50, trained the same way; \mbox{DenseNet-121}'s transition pooling rounds up, which prevents its feature map from vanishing at this resolution.

\subsection{LiteMed-ViT}
\begin{figure}[htbp]
    \centering
    \includegraphics[width=\columnwidth]{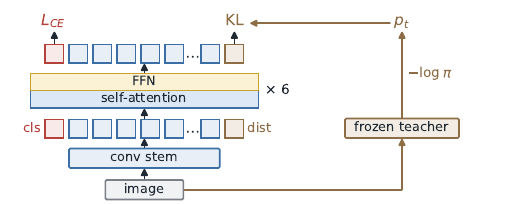}
    \caption{LiteMed-ViT training. The $cls$ output is trained with the logit-adjusted cross-entropy $L_{CE}$, and the $dist$ output by the KL divergence to $p_t$, the frozen teacher's output on the same image with the class prior removed (Eq.~\ref{eq:total_loss}). At inference the two outputs are averaged.}
    \label{fig:architecture}
\end{figure}

LiteMed-ViT (Fig.~\ref{fig:architecture}) replaces the patch embedding of a ViT~\cite{dosovitskiy_vit} with a convolutional stem, which eases the training of small transformers~\cite{xiao_earlyconv}: two stride-2 $3\times3$ convolutions (32 and 64 channels), each followed by group normalisation~\cite{wu_groupnorm}, give a $7\times7$ grid of 49 tokens of dimension 64; unlike batch normalisation, group normalisation does not depend on running statistics that corruption shifts. A depthwise $3\times3$ convolution adds a conditional positional encoding~\cite{chu_cpe}; $cls$ and $dist$ tokens~\cite{touvron_deit} and a learned positional embedding are added, and six pre-norm blocks follow, each with four-head self-attention and a feed-forward layer (width 128) that applies a depthwise $3\times3$ convolution to the patch tokens~\cite{yuan_ceit}, with LayerScale~\cite{touvron_cait} (initialised at 0.1) and stochastic depth~\cite{huang_stochdepth} rising to 0.1. The model has 0.234M parameters and 11.6M multiply--accumulates per image. It was selected from nine variants (token count, stem normalisation, class balancing, augmentation, small-data attention), each trained once per subset, on validation macro-F1; validation balanced accuracy decided between the two leaders, 0.2 points apart.

On DermaMNIST a classifier that minimises the average error favours the majority class at the expense of the rare ones. The student is therefore trained with logit adjustment~\cite{menon_logit} on every subset. With $\pi$ the training-set class frequencies, the $cls$ token is supervised on $z_{cls} + \log\pi$, and the $dist$ token matches a teacher target from which the same prior is removed. With $z_t$ the teacher logits, $T$ the temperature, $p_t=\sigma((z_t-\log\pi)/T)$ the prior-corrected teacher distribution and $p_s=\sigma(z_{dist}/T)$ the student's, the objective is \eqref{eq:total_loss},
\begin{equation}
L = L_{CE}(z_{cls} + \log\pi,\, y) + \alpha\, T^2\, \mathrm{KL}(p_t \,\|\, p_s),
\label{eq:total_loss}
\end{equation}
with $T=4$ and $\alpha=0.5$. At inference the raw logits are averaged, $(z_{cls}+z_{dist})/2$; because the prior enters only during training, this prediction targets balanced accuracy, and adding $\log\pi$ back recovers the standard prediction.

All models in the main comparison are trained for 50 epochs with AdamW (learning rate $3\times10^{-4}$, weight decay $10^{-4}$), cosine decay and batch size 128. The student is trained from scratch with dropout $p=0.1$ on every subset; the CNN baselines, MobileNetV3-Small, SqueezeNet~1.1 and ShuffleNet~V2 ($0.5\times$), are ImageNet-pretrained and fine-tuned at $28\times28$ with the same budget. Each run retains the checkpoint with the best validation accuracy (for the student, under its balanced rule).

\subsection{Evaluation Protocol}
\label{sec:protocol}
Every model in the main comparison is trained with four seeds, and the mean with a 95\% $t$-interval is reported. Under the pre-set rule, two models are said to differ only when a Welch test gives $p<0.05$ \emph{and} the mean difference exceeds 1.05 points, the change observed on retraining an identical configuration in preliminary experiments; otherwise they are described as comparable. Because only LiteMed-ViT is trained with class balancing, its raw accuracy is not compared with a baseline's: models are compared under a matched decision rule, either all standard (LiteMed-ViT's logits plus $\log\pi$) or all balanced (the baselines' logits minus $\log\pi$, the post-hoc form of logit adjustment~\cite{menon_logit}), both computed from the saved outputs of the same trained models. The teacher study and the hyperparameter sweeps (one seed per setting, four at the default) use a shorter protocol (25 epochs; PathMNIST subsampled to a stratified 15{,}000 images), and the resolution study retrains LiteMed-ViT and a ResNet18 on DermaMNIST at 28, 64, 128 and 224\,px (MedMNIST+~\cite{doerrich_medmnistplus}) for 30 epochs; each study is compared only within itself.

\section{Results and Discussion}

\begin{table*}[t]
\caption{Efficiency and test performance, mean\,$\pm$\,95\% confidence interval over four seeds: accuracy (Acc) under the standard and balanced accuracy (bAcc) under the balanced decision rule for every model (Section~\ref{sec:protocol}); latency at batch size one on a four-core CPU and a T4 GPU. $^{\uparrow}$/$^{\downarrow}$: higher/lower than LiteMed-ViT under the pre-set rule; unmarked: comparable. Bold: best compact model. ResNet18, the teacher, is shown for reference.}
\label{tab:comprehensive_comparison}
\centering
\footnotesize
\setlength{\tabcolsep}{3.2pt}
\begin{tabular}{@{}lcccc cc cc cc@{}}
\toprule
\multirow{2}{*}{\textbf{Model}} & \textbf{Params} & \textbf{MACs} & \textbf{CPU} & \textbf{GPU} & \multicolumn{2}{c}{\textbf{BloodMNIST}} & \multicolumn{2}{c}{\textbf{DermaMNIST}} & \multicolumn{2}{c}{\textbf{PathMNIST}} \\
\cmidrule(lr){6-7}\cmidrule(lr){8-9}\cmidrule(lr){10-11}
 & \textbf{(M)} & \textbf{(M)} & \textbf{(ms)} & \textbf{(ms)} & Acc & bAcc & Acc & bAcc & Acc & bAcc \\
\midrule
\textit{ResNet18} & 11.181 & 34.41 & 6.73 & 2.38 & 96.29{\scriptsize$\pm$0.15}$^{\uparrow}$ & 96.16{\scriptsize$\pm$0.22}$^{\uparrow}$ & 76.95{\scriptsize$\pm$2.29}$^{\uparrow}$ & 59.86{\scriptsize$\pm$2.16}$^{\uparrow}$ & 90.63{\scriptsize$\pm$0.32}$^{\uparrow}$ & 88.12{\scriptsize$\pm$0.60}$^{\uparrow}$ \\
\midrule
MobileNetV3-Small & 1.526 & 2.23 & 7.79 & 5.58 & 91.68{\scriptsize$\pm$0.61}$^{\downarrow}$ & 90.66{\scriptsize$\pm$0.70}$^{\downarrow}$ & 74.11{\scriptsize$\pm$0.56} & 56.62{\scriptsize$\pm$3.24} & 84.61{\scriptsize$\pm$0.38}$^{\downarrow}$ & 80.77{\scriptsize$\pm$0.60}$^{\downarrow}$ \\
SqueezeNet 1.1 & 0.727 & 2.59 & 1.81 & 1.97 & 93.91{\scriptsize$\pm$0.63}$^{\downarrow}$ & 93.36{\scriptsize$\pm$0.86}$^{\downarrow}$ & \textbf{75.62}{\scriptsize$\pm$0.70}$^{\uparrow}$ & 55.60{\scriptsize$\pm$1.45} & 88.39{\scriptsize$\pm$0.62} & 85.20{\scriptsize$\pm$0.64} \\
ShuffleNet V2 0.5$\times$ & 0.350 & 0.84 & 10.45 & 7.45 & 92.96{\scriptsize$\pm$0.56}$^{\downarrow}$ & 92.22{\scriptsize$\pm$0.45}$^{\downarrow}$ & 75.45{\scriptsize$\pm$0.81}$^{\uparrow}$ & \textbf{57.67}{\scriptsize$\pm$0.90}$^{\uparrow}$ & 86.65{\scriptsize$\pm$0.39}$^{\downarrow}$ & 83.30{\scriptsize$\pm$0.64}$^{\downarrow}$ \\
\textbf{LiteMed-ViT} & 0.234 & 11.64 & 7.23 & 4.42 & \textbf{95.05}{\scriptsize$\pm$0.55} & \textbf{94.48}{\scriptsize$\pm$0.59} & 74.26{\scriptsize$\pm$0.53} & 55.73{\scriptsize$\pm$1.98} & \textbf{89.15}{\scriptsize$\pm$0.93} & \textbf{85.56}{\scriptsize$\pm$1.00} \\
\bottomrule
\end{tabular}

\end{table*}

\begin{figure*}[t]
    \centering
    \includegraphics[width=\textwidth]{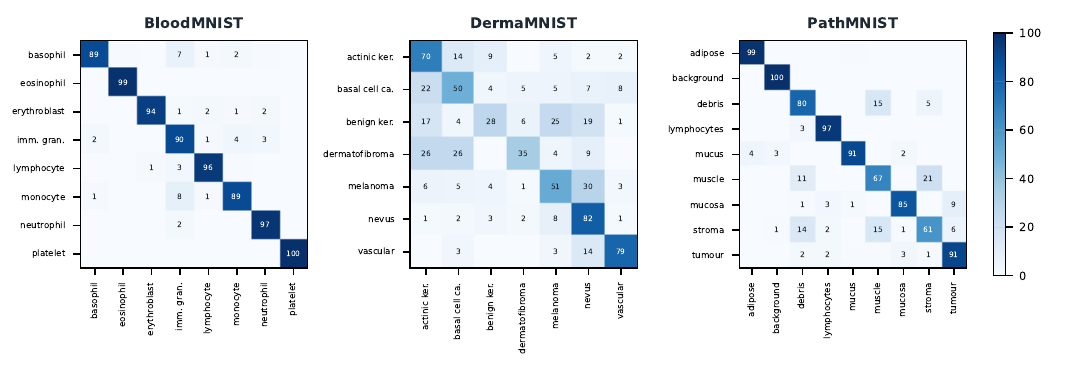}
    \caption{Confusion matrices of LiteMed-ViT (seed 0, balanced decision rule) on the full BloodMNIST, DermaMNIST and PathMNIST test sets. Rows are true classes and cells are row-normalised percentages; class names are abbreviated.}
    \label{fig:confusion_matrix}
\end{figure*}

LiteMed-ViT is the smallest model evaluated (0.234M parameters and 0.89\,MB in fp32, $47.8\times$ fewer parameters than ResNet18) but not the cheapest to run (Table~\ref{tab:comprehensive_comparison}): its 49 tokens cost 11.64M multiply--accumulates per image, $4.5$--$13.8\times$ more than the compact CNNs, and at batch size one it takes 7.23\,ms on the CPU and 4.42\,ms on the GPU, less than MobileNetV3-Small and ShuffleNet~V2 but more than SqueezeNet~1.1 (1.81 and 1.97\,ms) and ResNet18 (6.73 and 2.38\,ms).

On BloodMNIST LiteMed-ViT is more accurate than every compact CNN under both rules: $95.05\%$ accuracy against $93.91\%$ for the strongest, SqueezeNet~1.1, and $94.48\%$ balanced accuracy against $93.36\%$. Both margins satisfy the pre-set rule, although narrowly (1.1 points each), while ResNet18 is 1.2 and 1.7 points higher. On PathMNIST its $89.15\%$ accuracy and $85.56\%$ balanced accuracy exceed those of MobileNetV3-Small and ShuffleNet~V2 but are comparable to SqueezeNet~1.1's ($88.39\%$ and $85.20\%$; $p=0.08$ and $0.37$), while ResNet18 is 1.5 and 2.6 points higher.

DermaMNIST exhibits a different trend. LiteMed-ViT is comparable to MobileNetV3-Small and 1.2--1.4 points below ShuffleNet~V2 and SqueezeNet~1.1 in accuracy, and comparable to MobileNetV3-Small and SqueezeNet~1.1 in balanced accuracy, 1.9 points below ShuffleNet~V2; ResNet18 is higher on both. Switching to the balanced rule raises LiteMed-ViT’s balanced accuracy by 13.2 points while reducing its standard accuracy by 4.8 points. Table~\ref{tab:derma_perclass} shows the same trained LiteMed-ViT under both rules: under the standard rule it never predicts dermatofibroma (recall 0) and recalls only $0.31$ of melanomas; under the balanced rule these rise to $0.29$ and $0.56$, while nevus recall falls from $0.95$ to $0.80$. Minority-class underperformance is therefore a property of the decision rule, shared by all evaluated models, rather than of the transformer, and logit adjustment removes most of it.

\begin{figure}[htbp]
    \centering
    \includegraphics[width=\columnwidth]{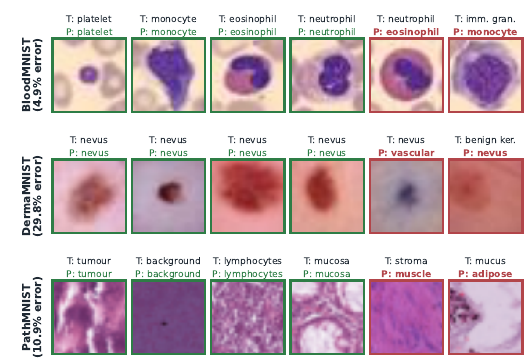}
    \caption{LiteMed-ViT test predictions (seed 0, balanced rule): per subset, four random correct predictions and the two most confident errors (highest probability on a wrong class). \texttt{T:} truth, \texttt{P:} prediction. Row labels give the model's test error rate on each subset.}
    \label{fig:predictions}
\end{figure}

The confusion matrices (Fig.~\ref{fig:confusion_matrix}) show where the remaining errors fall. On BloodMNIST most errors go to immature granulocytes. On PathMNIST they are concentrated among smooth muscle, cancer-associated stroma and debris. On DermaMNIST benign keratosis is split between melanoma and nevus, and dermatofibroma between actinic keratosis and basal cell carcinoma. Fig.~\ref{fig:predictions} shows, for each subset, LiteMed-ViT's most confident errors beside correct predictions.

\begin{table}[htbp]
    \caption{Per-class DermaMNIST test results of LiteMed-ViT (mean of four seeds): the same trained models under the standard and balanced rules.The last row gives the total support and the means of the metrics; mean recall is the balanced accuracy.}
    \label{tab:derma_perclass}
    \centering
    \footnotesize
    \setlength{\tabcolsep}{4pt}
    \begin{tabular}{@{}lr c ccc@{}}
\toprule
 & & \textbf{Standard} & \multicolumn{3}{c}{\textbf{Balanced}} \\
\cmidrule(lr){3-3}\cmidrule(lr){4-6}
\textbf{Class} & \textbf{Supp.} & Recall & Precision & Recall & F1 \\
\midrule
Dermatofibroma & 23 & 0.000 & 0.152 & 0.293 & 0.200 \\
Vascular lesion & 29 & 0.638 & 0.401 & 0.819 & 0.533 \\
Actinic keratosis & 66 & 0.356 & 0.325 & 0.640 & 0.431 \\
Basal cell carcinoma & 103 & 0.447 & 0.481 & 0.447 & 0.461 \\
Benign keratosis & 220 & 0.283 & 0.462 & 0.340 & 0.389 \\
Melanoma & 223 & 0.306 & 0.379 & 0.562 & 0.451 \\
Melanocytic nevus & 1341 & 0.947 & 0.911 & 0.801 & 0.852 \\
\midrule
\textit{Mean / total} & 
\textit{2005}& \textit{0.425} & \textit{0.445} & \textit{0.557} & \textit{0.474} \\
\bottomrule
\end{tabular}

\end{table}

\begin{table*}[t]
    \caption{Accuracy (\%) under \mbox{MedMNIST-C} corruptions at severities 1, 3 and 5, and mean corrupted accuracy (mCA) over all corruptions and severities; mean of four seeds, standard decision rule. $^{\uparrow}$/$^{\downarrow}$: mCA higher/lower than LiteMed-ViT's under the pre-set rule. Bold: best compact model. Mean: mCA averaged over the three subsets.}
    \label{tab:robustness}
    \centering
    \footnotesize
    \setlength{\tabcolsep}{4pt}
\begin{tabular}{@{}lcccc@{\hspace{8pt}}cccc@{\hspace{8pt}}cccc@{\hspace{8pt}}c@{}}
\toprule
 & \multicolumn{4}{c}{\textbf{BloodMNIST}} & \multicolumn{4}{c}{\textbf{DermaMNIST}} & \multicolumn{4}{c}{\textbf{PathMNIST}} & \\
\cmidrule(r){2-5}\cmidrule(r){6-9}\cmidrule(r){10-13}
\textbf{Model} & s1 & s3 & s5 & mCA & s1 & s3 & s5 & mCA & s1 & s3 & s5 & mCA & \textbf{Mean} \\
\midrule
\textit{ResNet18} & 89.59 & 69.72 & 58.81 & 72.47$^{\downarrow}$ & 75.83 & 72.71 & 68.39 & 72.35 & 85.79 & 80.44 & 70.15 & 79.11$^{\uparrow}$ & 74.64 \\
\midrule
MobileNetV3-Small & 81.50 & 64.00 & 54.88 & 66.00$^{\downarrow}$ & 72.92 & 69.78 & 64.48 & 69.18$^{\downarrow}$ & 80.92 & 74.66 & 62.83 & 73.25$^{\downarrow}$ & 69.48 \\
SqueezeNet 1.1 & 87.08 & 71.49 & 59.78 & 72.78$^{\downarrow}$ & \textbf{74.82} & 71.45 & 65.28 & 70.75 & 81.63 & 76.64 & 66.69 & 75.27$^{\downarrow}$ & 72.93 \\
ShuffleNet V2 0.5$\times$ & 83.88 & 62.61 & 55.07 & 66.30$^{\downarrow}$ & 74.13 & 70.94 & 65.35 & 70.28$^{\downarrow}$ & 81.72 & 76.78 & 66.59 & 75.34$^{\downarrow}$ & 70.64 \\
\textbf{LiteMed-ViT} & \textbf{91.35} & \textbf{72.95} & \textbf{63.01} & \textbf{75.59} & 73.81 & \textbf{72.50} & \textbf{69.08} & \textbf{71.70} & \textbf{84.84} & \textbf{79.49} & \textbf{68.38} & \textbf{77.99} & \textbf{75.09} \\
\bottomrule
\end{tabular}

\end{table*}

\begin{table*}[t]
    \caption{Teacher study (shorter protocol, Section~\ref{sec:protocol}): accuracy (\%) of each teacher (T, one run) and of the LiteMed-ViT student distilled from it (S), standard rule, and the student's balanced accuracy (S$_b$), balanced rule; students are mean\,$\pm$\,95\% CI over four seeds. $^{\uparrow}$/$^{\downarrow}$: higher/lower than the ResNet18-taught student under the pre-set rule; unmarked: comparable.}
    \label{tab:teachers}
    \centering
    \footnotesize
    \setlength{\tabcolsep}{4pt}
    \begin{tabular}{@{}lccccccccc@{}}
\toprule
 & \multicolumn{3}{c}{\textbf{BloodMNIST}} & \multicolumn{3}{c}{\textbf{DermaMNIST}} & \multicolumn{3}{c}{\textbf{PathMNIST}} \\
\cmidrule(lr){2-4}\cmidrule(lr){5-7}\cmidrule(lr){8-10}
\textbf{Teacher} & T & S & S$_b$ & T & S & S$_b$ & T & S & S$_b$ \\
\midrule
ResNet18 & 96.4 & 93.7{\scriptsize$\pm$0.2} & 92.8{\scriptsize$\pm$0.6} & 77.0 & 72.7{\scriptsize$\pm$1.6} & 51.9{\scriptsize$\pm$3.0} & 87.9 & 85.4{\scriptsize$\pm$1.1} & 80.6{\scriptsize$\pm$0.8} \\
ResNet50 & 96.6 & 93.6{\scriptsize$\pm$0.2} & 92.6{\scriptsize$\pm$0.5} & 79.8 & 71.7{\scriptsize$\pm$2.0} & 42.6{\scriptsize$\pm$5.7}$^{\downarrow}$ & 89.5 & 85.8{\scriptsize$\pm$0.9} & 80.7{\scriptsize$\pm$1.1} \\
DenseNet-121 & 96.8 & 93.9{\scriptsize$\pm$0.4} & 93.1{\scriptsize$\pm$0.4} & 79.4 & 71.9{\scriptsize$\pm$2.0} & 46.3{\scriptsize$\pm$4.9}$^{\downarrow}$ & 89.1 & 85.5{\scriptsize$\pm$1.1} & 80.5{\scriptsize$\pm$1.3} \\
Ens.\ R18+R50 & 97.1 & 93.7{\scriptsize$\pm$0.3} & 92.9{\scriptsize$\pm$0.3} & 80.4 & 71.9{\scriptsize$\pm$2.5} & 47.0{\scriptsize$\pm$5.4} & 90.1 & 85.6{\scriptsize$\pm$1.0} & 80.7{\scriptsize$\pm$1.0} \\
\bottomrule
\end{tabular}

\end{table*}

Under \mbox{MedMNIST-C} corruptions (Table~\ref{tab:robustness}, standard rule), LiteMed-ViT's BloodMNIST mCA of $75.59\%$ is above those of every compact CNN and of ResNet18 ($72.47\%$), and it is the most accurate of the five models at every severity. On DermaMNIST it is comparable to ResNet18 and SqueezeNet~1.1 and more robust than MobileNetV3-Small and ShuffleNet~V2. At severity~1 it is comparable to all four CNNs, but its accuracy declines least as severity increases, by 4.7 points from severity~1 to~5 against 7.4--9.5 for the others; at severity~5 it is comparable to ResNet18 and more accurate than every compact CNN, whose accuracy falls below the $66.9\%$ obtained by always predicting nevus. On PathMNIST its mCA of $77.99\%$ is above every compact CNN and it is the most accurate compact model at every severity; ResNet18 is higher but retains a similar share of its clean accuracy ($87.3\%$ against $87.5\%$). Under the balanced rule its mCA is above every compact CNN's on all three subsets, including SqueezeNet~1.1 on DermaMNIST ($64.01\%$ against $59.97\%$).

In the teacher study (Table~\ref{tab:teachers}), replacing ResNet18 with ResNet50, \mbox{DenseNet-121} or the ResNet18+ResNet50 ensemble raises teacher accuracy by up to 0.7, 3.4 and 2.2 points on BloodMNIST, DermaMNIST and PathMNIST, but no stronger teacher produces a more accurate student: every accuracy difference is within 1.0 point and none passes the pre-set rule, as reported for larger teachers~\cite{cho_efficacy}. At $28\times28$ the accuracy plateau is therefore the student's, not the teacher's. On DermaMNIST, ResNet50 and \mbox{DenseNet-121} reduce the student's balanced accuracy, from $51.9\%$ with ResNet18 to $42.6\%$ and $46.3\%$, both below it under the pre-set rule. The prior correction in \eqref{eq:total_loss} accounts for this: the same $\log\pi$ is subtracted from every teacher's logits, but the two are overconfident (expected calibration error 16.1\% and 12.9\%, against 6.0\% for ResNet18); consequently, less of their majority-class bias is removed, and the corrected target at $T=4$ places 0.46 and 0.34 of its mass on nevus, against 0.26. Temperature scaling of both teachers on the validation split~\cite{guo_calibration} confirms this: it reduces that mass to 0.17 and raises their students' balanced accuracy to $49.9\%$ and $51.1\%$, above the uncalibrated students' under the pre-set rule and comparable to the ResNet18-taught student's, at unchanged accuracy. On the other two subsets the correction is negligible and every teacher yields a comparable student; with class balancing, a teacher's calibration matters as well as its accuracy.

The distillation settings have little effect: varying $T\in\{1,2,8\}$ or $\alpha\in\{0.1,0.3,0.7,0.9\}$ one at a time leaves the student's validation macro-F1 within two standard deviations of the four-seed default on all three subsets, although $\alpha$ also rescales the loss, the cross-entropy weight being fixed at one. Dropout $p=0.3$ falls below that band on BloodMNIST (macro-F1 $90.76\%$ against $93.78\%$) and PathMNIST ($79.67\%$ against $83.17\%$), while $p=0$ and $p=0.2$ remain within it, supporting $p=0.1$.

To test whether the 28-pixel inputs limit these conclusions, LiteMed-ViT and ResNet18 were retrained on DermaMNIST at 28 to 224\,px under this study's own 30-epoch protocol (Section~\ref{sec:protocol}; one or two seeds per size), and its 28-px results therefore differ from the 50-epoch results of Table~\ref{tab:comprehensive_comparison}. ResNet18 gains 12.6 points ($77.38\%$ to $89.98\%$), whereas LiteMed-ViT stays between $71.7\%$ and $73.6\%$ at every size, and the gap grows from 4.7 to 16.4 points. Two runs of the same model and size differ by up to 1.6 points: ResNet18's gain is far larger, whereas LiteMed-ViT's variation across sizes is comparable. Higher resolution is also costlier for LiteMed-ViT: its token count grows from 49 to 3{,}136, and the cost of self-attention grows with the square of that count.

Three limitations remain. First, the main comparison uses $28\times28$ inputs; the resolution study shows that higher resolutions benefit ResNet18 significantly ($89.98\%$ at 224 px, aligning with ~\cite{kumar_derma}) more than LiteMed-ViT. Second, because the corrupted test sets are generated at $128\times128$ using the benchmark's reference implementation, mCA evaluates models with one another rather than benchmarking against published \mbox{MedMNIST-C} results. Third, latency was benchmarked on a cloud hardware (CPU and T4 GPU) instead of edge devices.

\FloatBarrier
\section{Conclusion}

Deploying deep models for point-of-care medical analysis requires balancing compactness with performance. At $0.234\text{M}$ parameters and a $0.89\text{ MB}$ footprint, LiteMed-ViT outperforms compact CNNs on BloodMNIST and most on PathMNIST, demonstrates superior or comparable robustness under MedMNIST-C corruptions, and achieves compact-CNN accuracy at sub-megabyte size. However, it trades away inference speed, and logit adjustment is essential to mitigate minority-class failure on imbalanced dataset like DermaMNIST. Two broader lessons emerge. First, neither stronger teachers nor higher input resolutions improved student accuracy; for ultra-light models, performance is bound by its core training objective and architectural design rather than external scaling, effort is better spent on what it optimises than on what it is fed. Second, neither parameter count nor MACs ordered these models by measured latency (ResNet18 stores $48\times$ more weights yet runs faster on both processors), hence, a compactness claim should name the resource it means and rest on measurement. Larger students, resolution scaling on native data, and embedded hardware deployments remain open for future work.


\begin{thebibliography}{00}
\bibitem{litjens_survey} G. Litjens, T. Kooi, B. E. Bejnordi, A. A. A. Setio, F. Ciompi, M. Ghafoorian, J. A. van der Laak, B. van Ginneken, and C. I. S\'anchez, ``A survey on deep learning in medical image analysis,'' \textit{Medical Image Analysis}, vol. 42, pp. 60--88, 2017, doi: 10.1016/j.media.2017.07.005.
\bibitem{rancea_edge} A. Rancea, I. Anghel, and T. Cioara, ``Edge computing in healthcare: innovations, opportunities, and challenges,'' \textit{Future Internet}, vol. 16, no. 9, p. 329, 2024, doi: 10.3390/fi16090329.
\bibitem{dosovitskiy_vit} A. Dosovitskiy, L. Beyer, A. Kolesnikov, D. Weissenborn, X. Zhai, T. Unterthiner, M. Dehghani, M. Minderer, G. Heigold, S. Gelly, J. Uszkoreit, and N. Houlsby, ``An image is worth 16x16 words: transformers for image recognition at scale,'' in \textit{Int. Conf. on Learning Representations (ICLR)}, 2021, doi: 10.48550/arXiv.2010.11929.
\bibitem{hinton_kd} G. Hinton, O. Vinyals, and J. Dean, ``Distilling the knowledge in a neural network,'' in \textit{NIPS Deep Learning and Representation Learning Workshop}, 2014, doi: 10.48550/arXiv.1503.02531.
\bibitem{touvron_deit} H. Touvron, M. Cord, M. Douze, F. Massa, A. Sablayrolles, and H. J\'egou, ``Training data-efficient image transformers \& distillation through attention,'' in \textit{Int. Conf. on Machine Learning (ICML)}, 2021, pp. 10347--10357, doi: 10.48550/arXiv.2012.12877.
\bibitem{yang_medmnist} J. Yang, R. Shi, D. Wei, Z. Liu, L. Zhao, B. Ke, H. Pfister, and B. Ni, ``MedMNIST v2 -- a large-scale lightweight benchmark for 2D and 3D biomedical image classification,'' \textit{Scientific Data}, vol. 10, no. 1, p. 41, 2023, doi: 10.1038/s41597-022-01721-8.
\bibitem{iandola_squeezenet} F. N. Iandola, S. Han, M. W. Moskewicz, K. Ashraf, W. J. Dally, and K. Keutzer, ``SqueezeNet: AlexNet-level accuracy with 50x fewer parameters and $<$0.5MB model size,'' \textit{arXiv preprint arXiv:1602.07360}, 2016, doi: 10.48550/arXiv.1602.07360.
\bibitem{howard_mobilenetv3} A. Howard, M. Sandler, G. Chu, L.-C. Chen, B. Chen, M. Tan, W. Wang, Y. Zhu, R. Pang, V. Vasudevan, Q. V. Le, and H. Adam, ``Searching for MobileNetV3,'' in \textit{IEEE/CVF Int. Conf. on Computer Vision (ICCV)}, 2019, pp. 1314--1324, doi: 10.1109/ICCV.2019.00140.
\bibitem{ma_shufflenet} N. Ma, X. Zhang, H.-T. Zheng, and J. Sun, ``ShuffleNet V2: practical guidelines for efficient CNN architecture design,'' in \textit{European Conf. on Computer Vision (ECCV)}, 2018, pp. 122--138, doi: 10.1007/978-3-030-01264-9\_8
\bibitem{mehta_mobilevit} S. Mehta and M. Rastegari, ``MobileViT: light-weight, general-purpose, and mobile-friendly vision transformer,'' in \textit{Int. Conf. on Learning Representations (ICLR)}, 2022, doi: 10.48550/arXiv.2110.02178.
\bibitem{vaswani_attention} A. Vaswani, N. Shazeer, N. Parmar, J. Uszkoreit, L. Jones, A. N. Gomez, \L. Kaiser, and I. Polosukhin, ``Attention is all you need,'' in \textit{Advances in Neural Information Processing Systems (NeurIPS)}, 2017, pp. 5998--6008, doi: 10.48550/arXiv.1706.03762.
\bibitem{liu_swin} Z. Liu, Y. Lin, Y. Cao, H. Hu, Y. Wei, Z. Zhang, S. Lin, and B. Guo, ``Swin transformer: hierarchical vision transformer using shifted windows,'' in \textit{IEEE/CVF Int. Conf. on Computer Vision (ICCV)}, 2021, pp. 9992--1000, doi: 10.1109/ICCV48922.2021.00986.
\bibitem{matsoukas_replace} C. Matsoukas, J. F. Haslum, M. S\"oderberg, and K. Smith, ``Is it time to replace CNNs with transformers for medical images?,'' in \textit{ICCV Workshop on Computer Vision for Automated Medical Diagnosis (CVAMD)}, 2021, doi: 10.48550/arXiv.2108.09038.
\bibitem{shamshad_survey} F. Shamshad, S. Khan, S. W. Zamir, M. H. Khan, M. Hayat, F. S. Khan, and H. Fu, ``Transformers in medical imaging: a survey,'' \textit{Medical Image Analysis}, vol. 88, p. 102802, 2023, doi: 10.48550/arXiv.2201.09873.
\bibitem{romero_fitnets} A. Romero, N. Ballas, S. E. Kahou, A. Chassang, C. Gatta, and Y. Bengio, ``FitNets: hints for thin deep nets,'' in \textit{Int. Conf. on Learning Representations (ICLR)}, 2015, doi: 10.48550/arXiv.1412.6550.
\bibitem{gou_survey} J. Gou, B. Yu, S. J. Maybank, and D. Tao, ``Knowledge distillation: a survey,'' \textit{International Journal of Computer Vision}, vol. 129, no. 6, pp. 1789--1819, 2021, doi: 10.48550/arXiv.2006.05525.
\bibitem{cho_efficacy} J. H. Cho and B. Hariharan, ``On the efficacy of knowledge distillation,'' in \textit{IEEE/CVF Int. Conf. on Computer Vision (ICCV)}, 2019, pp. 4793--4801, doi: 10.1109/ICCV.2019.00489.
\bibitem{alshehri_hybrid} O. M. Alshehri, A. Shaf, M. Irfan, M. M. Jalal, M. A. Altayar, M. H. Abu-Alghayth, H. Al Shmrany, T. Ali, T. A. Soomro, and G. A. Alkhathami, ``A hybrid CNN-transformer framework for normal blood cell classification: towards automated hematological analysis,'' \textit{Computer Modeling in Engineering \& Sciences}, vol. 144, no. 1, pp. 1165--1196, 2025, doi: 10.32604/cmes.2025.067150.
\bibitem{li_hycovit} O. A. Naghash, N. Ling, and X. Li, ``HyCoViT: hybrid convolution vision transformer with dynamic dropout for enhanced medical chest X-ray classification,'' \textit{IEEE Access}, vol. 13, pp. 112623--112641, 2025, doi: 10.1109/ACCESS.2025.3584065.
\bibitem{acevedo_blood} A. Acevedo, A. Merino, S. Alf\'erez, \'A. Molina, L. Bold\'u, and J. Rodellar, ``A dataset of microscopic peripheral blood cell images for development of automatic recognition systems,'' \textit{Data in Brief}, vol. 30, p. 105474, 2020, doi: 10.1016/j.dib.2020.105474.
\bibitem{tschandl_ham10000} P. Tschandl, C. Rosendahl, and H. Kittler, ``The HAM10000 dataset, a large collection of multi-source dermatoscopic images of common pigmented skin lesions,'' \textit{Scientific Data}, vol. 5, p. 180161, 2018, doi:10.1038/sdata.2018.161.
\bibitem{kather_crc} J. N. Kather, J. Krisam, P. Charoentong, T. Luedde, E. Herpel, C.-A. Weis, T. Gaiser, A. Marx, N. A. Valous, D. Ferber, L. Jansen, C. C. Reyes-Aldasoro, I. Z\"ornig, D. J\"ager, H. Brenner, J. Chang-Claude, M. Hoffmeister, and N. Halama, ``Predicting survival from colorectal cancer histology slides using deep learning: a retrospective multicenter study,'' \textit{PLOS Medicine}, vol. 16, no. 1, art. no. e1002730, 2019, doi: DOI: 10.1371/journal.pmed.1002730.



\bibitem{nachbar_abcd} F. Nachbar, W. Stolz, T. Merkle, A. B. Cognetta, T. Vogt, M. Landthaler, P. Bilek, O. Braun-Falco, and G. Plewig, ``The ABCD rule of dermatoscopy: high prospective value in the diagnosis of doubtful melanocytic skin lesions,'' \textit{Journal of the American Academy of Dermatology}, vol. 30, no. 4, pp. 551--559, doi: 10.1016/s0190\mbox{-}9622(94)70061\mbox{-}3.
\bibitem{perez_aug} F. Perez, C. Vasconcelos, S. Avila, and E. Valle, ``Data augmentation for skin lesion analysis,'' in \textit{ISIC Skin Image Analysis Workshop}, Springer, Cham, 2018, pp. 303--311, doi:10.1007/978-3-030-01201-4\_33.
\bibitem{tellez_stain} D. Tellez, G. Litjens, P. B\'andi, W. Bulten, J. M. Bokhorst, F. Ciompi, and J. van der Laak, ``Quantifying the effects of data augmentation and stain color augmentation in convolutional neural networks for computational pathology,'' \textit{Medical Image Analysis}, vol. 58, p. 101544, 2019, doi: 10.1016/j.media.2019.101544.
\bibitem{disalvo_medmnistc} F. Di Salvo, S. Doerrich, and C. Ledig, ``\mbox{MedMNIST-C}: comprehensive benchmark and improved classifier robustness by simulating realistic image corruptions,'' in \textit{MICCAI Workshop on Advancing Data Solutions in Medical Imaging AI (ADSMI)}, 2024, doi:10.48550/arXiv.2406.17536.
\bibitem{he_resnet} K. He, X. Zhang, S. Ren, and J. Sun, ``Deep residual learning for image recognition,'' in \textit{IEEE Conf. on Computer Vision and Pattern Recognition (CVPR)}, 2016, pp. 770--778, doi: 10.1109/cvpr.2016.90.
\bibitem{huang_densenet} G. Huang, Z. Liu, L. van der Maaten, and K. Q. Weinberger, ``Densely connected convolutional networks,'' in \textit{IEEE Conf. on Computer Vision and Pattern Recognition (CVPR)}, 2017, pp. 2261--2269, doi: 10.1109/CVPR.2017.243.
\bibitem{xiao_earlyconv} T. Xiao, M. Singh, E. Mintun, T. Darrell, P. Doll\'ar, and R. Girshick, ``Early convolutions help transformers see better,'' in \textit{Advances in Neural Information Processing Systems (NeurIPS)}, vol. 34, 2021, pp. 30392--30400, doi: 10.48550/arXiv.2106.14881.
\bibitem{wu_groupnorm} Y. Wu and K. He, ``Group normalization,'' in \textit{European Conf. on Computer Vision (ECCV)}, 2018, pp. 3--19, doi: 10.1007/978-3-030-01261-8\_1.
\bibitem{chu_cpe} X. Chu, Z. Tian, B. Zhang, X. Wang, and C. Shen, ``Conditional positional encodings for vision transformers,'' in \textit{Int. Conf. on Learning Representations (ICLR)}, 2023, doi: 10.48550/arXiv.2102.10882.
\bibitem{yuan_ceit} K. Yuan, S. Guo, Z. Liu, A. Zhou, F. Yu, and W. Wu, ``Incorporating convolution designs into visual transformers,'' in \textit{IEEE/CVF Int. Conf. on Computer Vision (ICCV)}, 2021, pp. 559--568, doi: 10.48550/arXiv.2103.11816.
\bibitem{touvron_cait} H. Touvron, M. Cord, A. Sablayrolles, G. Synnaeve, and H. J\'egou, ``Going deeper with image transformers,'' in \textit{IEEE/CVF Int. Conf. on Computer Vision (ICCV)}, 2021, pp. 32--42, doi: 10.1109/ICCV48922.2021.00010.
\bibitem{huang_stochdepth} G. Huang, Y. Sun, Z. Liu, D. Sedra, and K. Q. Weinberger, ``Deep networks with stochastic depth,'' in \textit{European Conf. on Computer Vision (ECCV)}, 2016, pp. 646--661, doi: 10.1007/978-3-319-46493-0\_39.
\bibitem{menon_logit} A. K. Menon, S. Jayasumana, A. S. Rawat, H. Jain, A. Veit, and S. Kumar, ``Long-tail learning via logit adjustment,'' in \textit{Int. Conf. on Learning Representations (ICLR)}, 2021, doi: 10.48550/arXiv.2007.07314.
\bibitem{doerrich_medmnistplus} S. Doerrich, F. Di Salvo, J. Brockmann, and C. Ledig, ``Rethinking model prototyping through the MedMNIST+ dataset collection,'' \textit{Scientific Reports}, vol. 15, art. no. 7669, 2025, doi: 10.1038/s41598-025-92156-9.

\bibitem{guo_calibration} C. Guo, G. Pleiss, Y. Sun, and K. Q. Weinberger, ``On calibration of modern neural networks,'' in \textit{Int. Conf. on Machine Learning (ICML)}, 2017, pp. 1321--1330, doi: 10.48550/arXiv.1706.04599.
\bibitem{kumar_derma} A. Kumar, A. Vishwakarma, V. Bajaj, and S. Mishra, ``Novel mixed domain hand-crafted features for skin disease recognition using multiheaded CNN,'' \textit{IEEE Transactions on Instrumentation and Measurement}, vol. 73, pp. 1--13, 2024, doi: 10.1109/TIM.2024.3370772.
\end{thebibliography}
\end{document}